\documentclass[lettersize,journal]{IEEEtran}
\usepackage{amsmath,amsfonts}
\usepackage{algorithm}
\usepackage{array}
\usepackage[caption=false,font=normalsize,labelfont=sf,textfont=sf]{subfig}
\usepackage{textcomp}
\usepackage{stfloats}
\usepackage{url}
\usepackage{verbatim}
\usepackage{graphicx}
\usepackage{cite}
\usepackage{booktabs}
\usepackage{multirow}
\usepackage{makecell}
\usepackage{algorithm}
\usepackage{algpseudocode}
\usepackage{tabularx}

\begin{document}

\title{ViFP: A Framework for Visual False Positive Detection to Enhance Reasoning Reliability in VLMs}

\author{Ben Zhang, Lulu Yu, Lei Gao, Quanjiang Guo, Jing Liu, Hui Gao
\thanks{Ben Zhang, Lulu Yu, Lei Gao, Quanjiang Guo, and Jing Liu are with the School of Computer Science and Engineering, University of Electronic Science and Technology of China, Chengdu 611731, China (E-mail: zhangben@std.uestc.edu.cn; yululu@std.uestc.edu.cn; leigao@std.uestc.edu.cn; guochance1999@163.com; liuj1017@163.com).

Hui Gao is with the School of Computer Science and Engineering, University of Electronic Science and Technology of China, Chengdu , China and also with Kash Institute of Electronics and Information Industry, Kash 844000, China (E-mail: huigao@uestc.edu.cn).}}

\markboth{IEEE TRANSACTIONS ON MULTIMEDIA,~Vol.~XX, No.~XX, XX~2025}%
{Shell \MakeLowercase{\textit{et al.}}: A Sample Article Using IEEEtran.cls for IEEE Journals}


\maketitle

\begin{abstract}
During reasoning in vision-language models (VLMs), false positive (FP) reasoning occurs when a model produces the correct answer but follows an incorrect reasoning path, resulting in undermined reasoning reliability. Existing approaches mainly rely on prompt engineering, knowledge distillation or reinforcement learning to improve reasoning reliability, both of which require large amounts of high-quality data and thus limit practical applicability. Few approaches have focused on directly detecting and correcting FPs. To address these issues, we propose ViFP, a framework for Visual False Positive Detection to Enhance Reasoning Reliability in VLMs. ViFP builds effective reasoning paths through multi-turn QA and dynamically analyzes the consistency of the reasoning path to identify potential FPs. It also introduces a targeted reasoning chain correction mechanism to modify FP reasoning, thereby improving logical consistency and accuracy. Finally, we introduce a reliability evaluation metric—VoC, which integrates answer accuracy and the FP rate, providing a quantitative tool to assess whether a VLM not only answers correctly but also reasons reliably. Our experiments on closed-source VLMs show that ViFP consistently improves performance across three datasets: A-OKVQA, OK-VQA, and FVQA. On A-OKVQA, ViFP improves accuracy by up to 5.4\%, surpassing the previous state-of-the-art by 4.3\%, and significantly reduces the number of FPs, validating its benefits in enhancing reasoning reliability.

\end{abstract}

\begin{IEEEkeywords}
Visual Reasoning, Vision-Language Models, False Positive Reasoning, Reasoning Reliability, Reliability Evaluation.
\end{IEEEkeywords}

\section{Introduction}
\begin{figure}[!t]
    \centering
    \includegraphics[width=\linewidth]{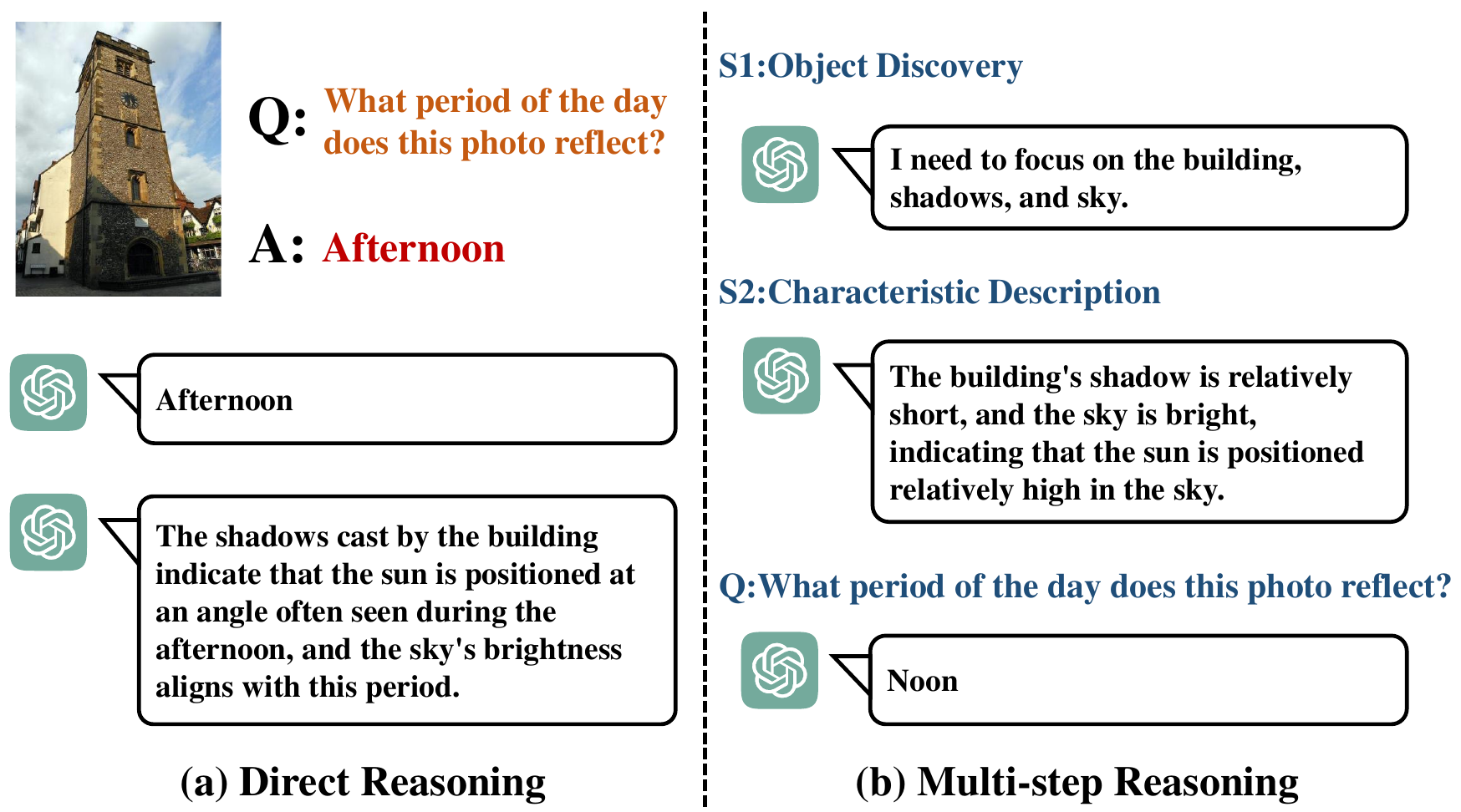}
    \caption{FP instance in visual reasoning. The VLM follows the same reasoning path twice (sun$\rightarrow$building$\rightarrow$shadows) but produces different answers (afternoon and noon). }
    \label{Fig1}
\end{figure}

Recent studies have shown that vision-language models (VLMs) achieve outstanding performance on visual question answering (VQA) \cite{li_blip-2_2023, hu_bliva_2024, jian_large_2024, yao_minicpm-v_2024, huang_mini-monkey_2024}. By integrating visual encoders with large language models (LLMs), VLMs jointly encode images and text into a unified vector space. This makes use of the powerful language understanding and generation capabilities of LLMs, while also enabling prompt engineering to generate structured reasoning paths. When the model’s output includes explicit reasoning chains, the phenomenon of false positive (FP) reasoning may emerge. An FP refers to instances where the model produces the correct final answer but follows an incorrect reasoning path \cite{wang_examining_2025}, as illustrated in Fig. \ref{Fig1}. A key cause of FPs lies in the model performing "illusory reasoning", whereby the reasoning path is not derived through step-by-step deduction, but instead reverse-engineered to justify a predetermined answer. Therefore, FPs are a critical factor that undermines the reliability of visual reasoning.

To improve reasoning reliability, some existing approaches use Chain-of-Thought (CoT) prompting to guide VLMs in generating explicit reasoning paths that support conclusions \cite{luo_improve_2024,setlur_rewarding_2024,wei_chain--thought_2022,trung_reft_2024}. Rather than explicitly identifying FP, these methods aim to reduce such errors by strengthening the model's logical reasoning capabilities. CoT decomposes complex questions into simpler sub-questions and forces the generation of intermediate reasoning paths \cite{tan_reason-rft_2025,chen_visual_2024,zhang_igniting_2025}. Some approaches use a standard template and assign the decomposition task to the VLM. This strategy does not provide any effective reasoning clues or supplementary information. The decomposition process is prone to deviations or errors. Distillation methods typically leverage powerful closed-source VLMs like GPT to generate high-quality training data, which is then used to train a smaller, more resource-efficient model, such as LLaVA-CoT \cite{shao_visual_2024,xu_llava-cot_2025}. However, due to the smaller model's inherent capacity limitations, the distilled model cannot surpass the reasoning capabilities of the original high-performing VLM.  While this approach enables computationally efficient, small-scale models to be distilled, it is not suitable for directly improving high-performing VLMs.

The above approaches aim to improve the reliability of reasoning through forward enhancement. However, few approaches address the issue of reverse verification, particularly with regard to detecting and correcting FP. Current FP detection methods fall into two categories: human and model-based \cite {wang_examining_2025}. The former is inefficient and lacks scalability, while the latter, although automated, typically relies on high-performance supervisory models. This creates a dilemma: if a superior VLM is needed to judge reasoning validity, \IEEEpubidadjcol it is often more efficient to use that superior VLM for reasoning. Meanwhile, both human and model-based evaluations suffer from inconsistent cognitive standards in FP. Assume a supervisory model to review the reasoning process of the inference model. As shown in the example in Fig. \ref{Fig1}, a supervisory model receiving only the direct reasoning (left) might deem it valid (e.g. accepting "reasoning time from sunlight and shadows" as feasible). However, if in reality the model cannot progress from analyzing the lighting to reasoning the time (right), it suggests that the reasoning logic is flawed and should be classified as a FP. This allows users to identify which parts of the VLM's reasoning are unreliable. By designing effective reasoning strategies or fine-tuning the model based on FP instances, the accuracy and reliability of VLM visual reasoning can be improved.

In summary, we propose ViFP, a framework for visual FP detection to enhance reasoning reliability in VLMs. ViFP is compatible with leading closed-source VLMs including  GPT-4o, Gemini-2.5, and Grok-4. ViFP employs a prompt-based evaluation to detect FPs by assessing the consistency between direct and multi-step reasoning outputs. Meanwhile, an exclusive correction mechanism has been introduced to optimize multi-step reasoning paths, improving overall reliability in terms of both answer accuracy and reasoning path reliability. ViFP constructs a generalizable sub-question set derived from real-world question-solving logic, and defines the mapping between question types and reasoning chains. This requires the model to perform targeted, multi-step reasoning based on the question type. This design ensures the normativity of the reasoning process, while enabling the model to adapt to complex scenarios by adjusting the reasoning chain flexibly.

Finally, we conducted experiments to evaluate ViFP on three real-world VQA datasets, demonstrating its ability to detect and correct FPs, as well as improve both the reasoning accuracy and reasoning path reliability of base VLMs. The main contributions of our method are as follows:
\begin{itemize}
    \item We propose ViFP, a fast and lightweight FP self-detection framework that can be seamlessly applied to various VLMs;
    
    \item We introduce a visual reasoning optimization method based on feedback from FPs. This enables VLMs to adapt the reasoning chain dynamically, achieving more accurate and reliable reasoning;
    
    \item We introduce a novel metric for assessing reasoning reliability — VoC(Value of Correction) and validate it on closed-source VLMs, demonstrating its preliminary feasibility in assessing reasoning reliability.
\end{itemize}

\section{Related Work}
\subsection{Visual Reasoning in VLMs} 
VQA tasks require models to possess visual perception, cognitive reasoning, and extensive knowledge resources \cite{johnson_clevr_2017,noauthor_review_2023}. With these three objectives in mind, researchers have continually explored more efficient and unified approaches. BLIP-2 introduces a Q-Former module that selectively attends to important regions in the image, enhancing accuracy but increasing hallucinations \cite{li_blip-2_2023}. MiniCPM and Mini-Monkey have designed special image patch segmentation strategies and reduced model size for deployment on the device side, resulting in weak multi-step inference capabilities \cite{yao_minicpm-v_2024, huang_mini-monkey_2024}. MoMe, Omni-SMoLA, and VLMo flexibly use Mixture-of-Experts (MoE) to adapt to various tasks and improve generalization, but this increases training costs \cite{shen_mome_2024,wu_omni-smola_2024,bao_vlmo_2022}. GraphVis and GeReA incorporate external knowledge graphs to assist reasoning and expand the model's knowledge base, but these methods rely on high-quality knowledge bases \cite{tan_paths-over-graph_2025, hao_self-bootstrapped_2024, wen_multimodal_2024, ma_gerea_2024, deng_graphvis_2024}. DEDR further employs an iterative knowledge distillation approach over external knowledge bases such as Wikipedia to capture high-quality knowledge and generates answers using the MM-FiD decoder \cite{salemi_symmetric_2023}. MM-Reasoner generates detailed image captions and extracts key objects that appear. This information is used to prompt the LLM to dynamically generate relevant knowledge, thereby avoiding reliance on a traditional static knowledge base retrieval \cite{khademi_mm-reasoner_2023}. LLMs inherently store vast amounts of knowledge, with larger models offering richer information. Prophet leverages this capability by prompting LLMs to generate auxiliary knowledge, guiding a base VQA model to better identify the correct answer among candidates \cite{cao_knowledge_2024, yu_prophet_2025}.


\subsection{Reliable Reasoning in VLMs} 
The reliability of reasoning has long been a central concern. Existing approaches primarily focus on enhancing the reasoning capabilities of models \cite{setlur_rewarding_2024,luo_improve_2024,trung_reft_2024}, often through CoT, which encourages the model to output intermediate steps and decompose complex questions into simpler sub-questions \cite{wei_chain--thought_2022}. For example, prompting strategies like "Let's think step by step." and LLaVA-CoT assign the task of question decomposition to the powerful VLMs by designing general-purpose question templates. While these approaches promote versatility, they overly rely on the model's inherent reasoning ability and fail to consistently improve the performance of these powerful VLMs \cite{shao_visual_2024,xu_llava-cot_2025}. Visual-CoT further builds upon the idea of question decomposition by annotating images with visual highlights of key regions essential for answering, effectively producing explicit visual attention \cite{shao_visual_2024}.

In addition, some recent works in mathematics and physics domains have attempted to detect FP reasoning by introducing specific benchmarks for FP identification, or by constructing targeted datasets to help models learn reliable reasoning logic \cite{wang_examining_2025,dong_seeing_2025,aldazharova_assessing_2024}.

\begin{figure*}[!t]
    \centering
    \includegraphics[width=\textwidth]{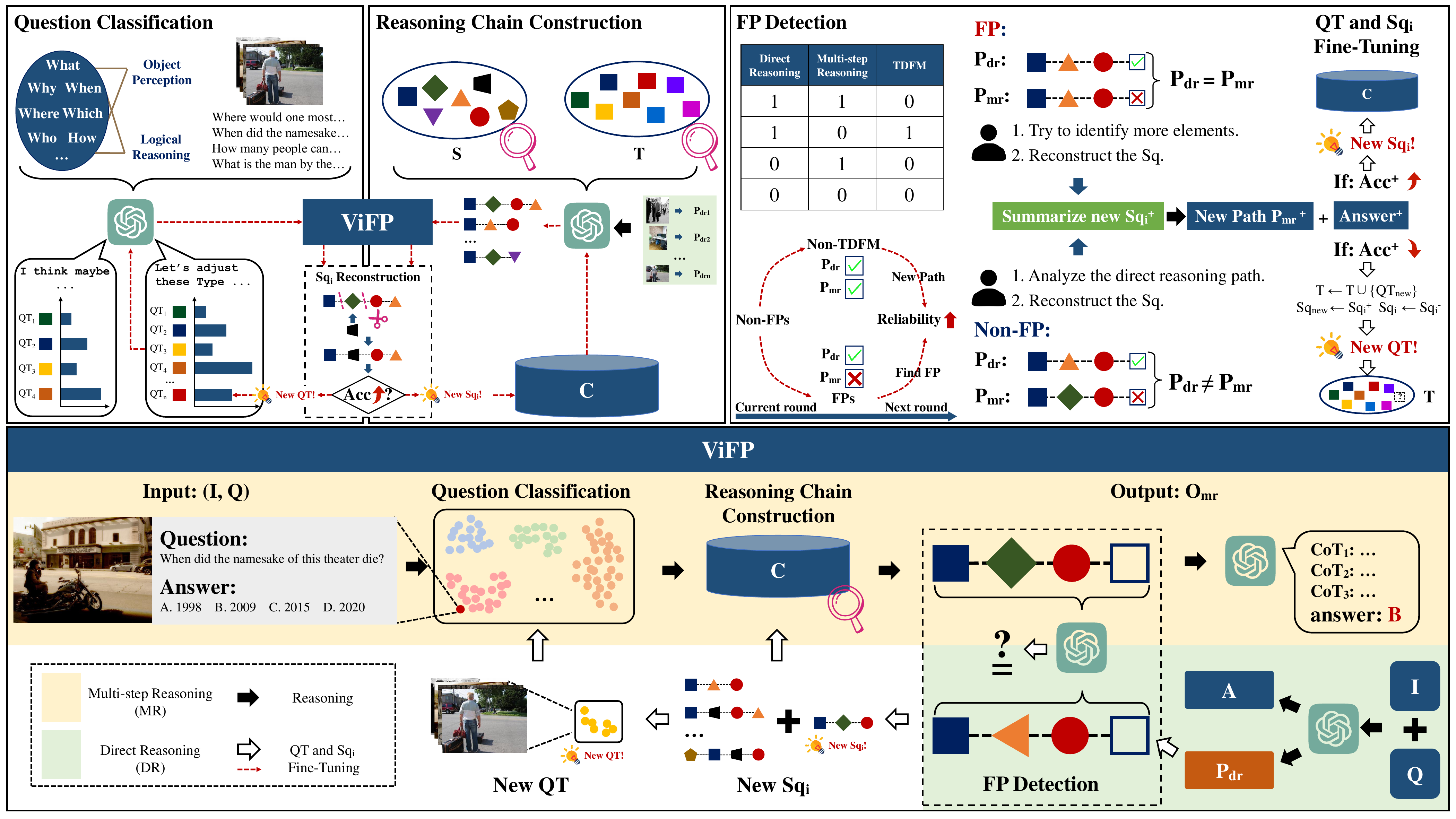}
    \caption{The overview framework of ViFP. 
    (1) Question Classification: real-world VQA questions are classified along the dimensions of perception and reasoning, forming the initial question types. 
    (2) Reasoning Chain Construction: the VLM summarizes reasoning chain templates for use in multi-step reasoning. 
    (3) FP Detection: detecting FPs by comparing the consistency between direct and multi-step reasoning. 
    (4) $QT$ and $Sq_i$ Fine-Tuning: Based on the detected FPs, the VLM refines the question types and the corresponding reasoning chains.
    (5) ViFP: Given an input (image question pair), ViFP conducts multi-step reasoning and iteratively refines its next reasoning to produce more accurate answers.}
    \label{Fig3}
\end{figure*}

\section{Methodology}
The overall architecture of ViFP is shown in Fig.\ref{Fig3}. Given an input image question pair, the VLM first generates a direct reasoning result. ViFP then classifies the question and constructs a corresponding reasoning chain composed of a sequence of sub-questions. The VLM sequentially answers the sub-questions, the outputs are used to detect FPs by examining the consistency with the direct reasoning. Finally, the detected FPs are used to discover new question types and optimize reasoning-chain templates, thereby improving the reliability of visual reasoning in VLM.



\subsection{Question Classification} 
The initial question types are generated by the VLM with the following prompt:

\begin{figure}[H]
    \centering
    \includegraphics[width=\linewidth]{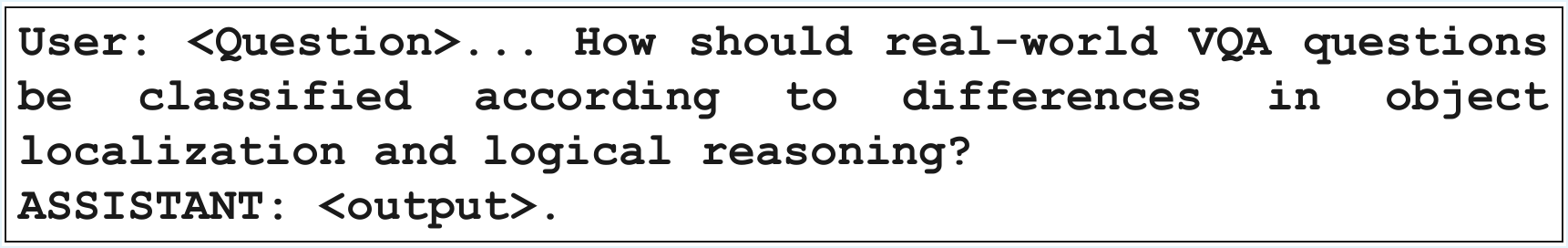}
    \label{Question Classification}
    \vspace{-10pt}
\end{figure}
The output is a set $T = \{QT_1, \dots, QT_k\}$, where the i-th question type is denoted as $QT_i$. “object localization” and “logical reasoning” are included in the prompt as few-shot cues to prevent the VLM from blindly initiating classification. This design can improve the reproducibility of ViFP.

\subsection{Reasoning Chain Construction}
First, the initial sub-question set is generated by the VLM with the following prompt:

\begin{figure}[H]
    \centering
    \includegraphics[width=\linewidth]{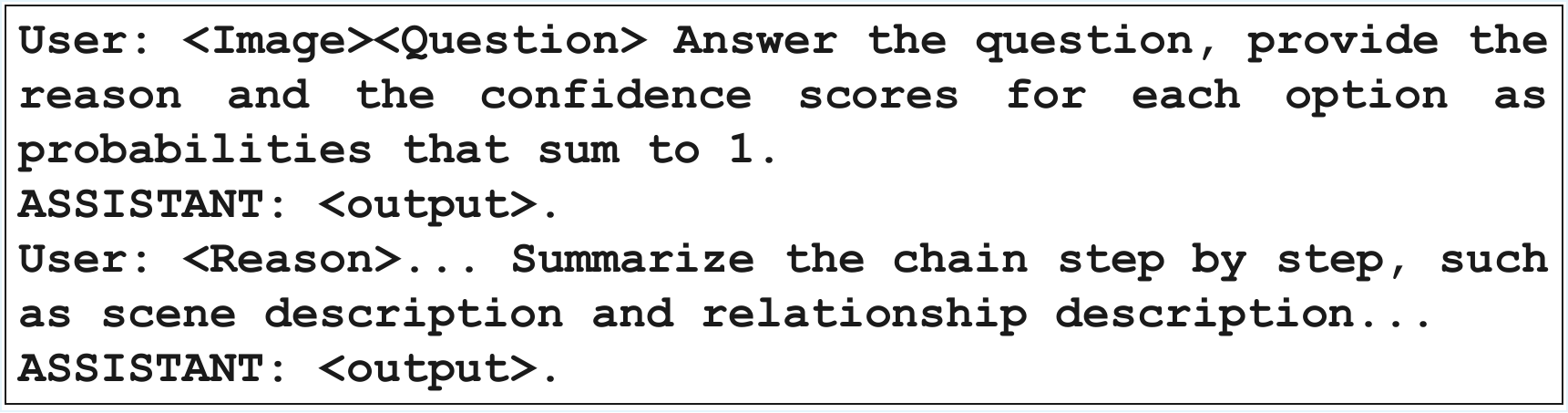}
    \label{Sub-question Generation}
    \vspace{-10pt}
\end{figure}
The first output is the direct reasoning result $O_{dr} = (P_{dr}, A)$, where $P$ denotes the reasoning path, the subscript “dr” stands for direct reasoning, and $A$ denotes the answer. Meanwhile, each option has a confidence score $p$ representing the probability of being the correct answer. The second output is a sequence of questions $Sq = \langle q_1, \ldots, q_n \rangle$, where $q$ denotes the sub-question.

Based on the reasoning path derived from direct reasoning, the VLM summarizes the reasoning chain template for each type of question by using the following prompt:
\begin{figure}[H]
    \centering
    \includegraphics[width=\linewidth]{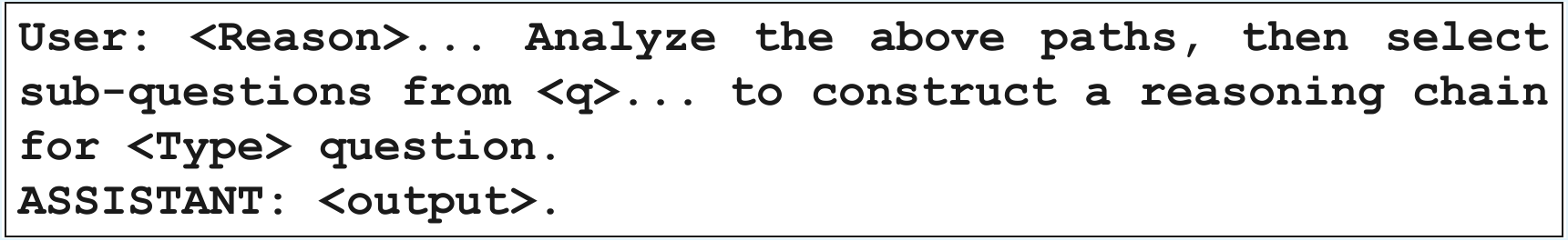}
    \label{Sq Construction}
    \vspace{-10pt}
\end{figure}
The output is the reasoning chain template $Sq_i = \{Sq \;\mid\; QT_i\}$, total number is $k$. 

In the next round of multi-step reasoning, $Sq_i$ is used as the reasoning chain, the output is $O_{mr} = (P_{mr}, A)$, where $P_{mr}$ is a sequence of QA pairs $P_{mr} = \langle(q_1, A_1), \dots, (q_n, A_n)\rangle$, and the subscript “mr” stands for multi-step reasoning.

\subsection{FP Detection} 
The error distributions differ between direct reasoning and multi-step reasoning. We define certain instances as TDFM\begin{equation}
\text{TDFM} = \{ (P_{dr}, P_{mr}) \mid f(P_{dr}) = A^{*}, \ f(P_{mr}) \neq A^{*} \},
\end{equation}
where $A^{*}$ denotes the ground truth, TFDMs are used for FP detection. To formally characterize, the visual reasoning process is defined as a mapping function $f (P) = A$.

Assume there are $(P_{dr},A_{dr})$ and $(P_{mr},A_{mr})$, if $P_{dr}=P_{mr}$, but $A_{dr} \neq A_{mr}$, it implies that $f (P) = A_{dr}$ and $f (P) = A_{mr}$ hold simultaneously, which violates the definition of a function. Therefore, such a mapping function cannot exist, the $P_{dr}$ is identified as a FP. ViFP modifies its $Sq$ in order to discover a more reliable path. The prompt used in this part is shown as follows:
\begin{figure}[H]
    \centering
    \includegraphics[width=\linewidth]{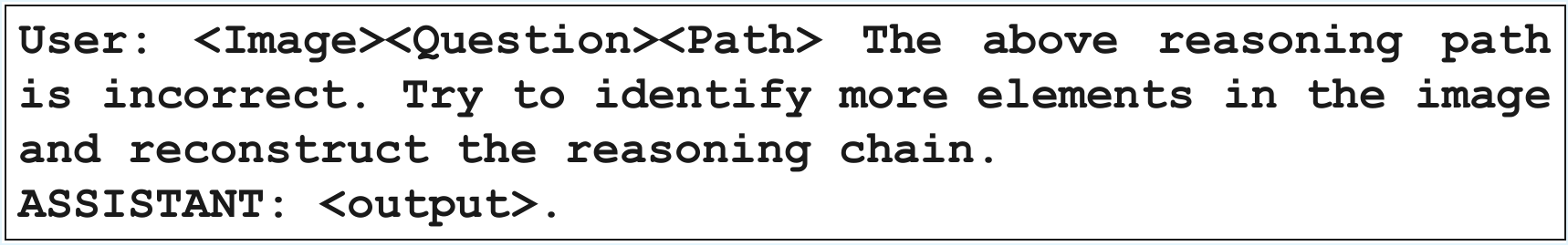}
    \label{FP Detection1}
    \vspace{-10pt}
\end{figure}
If $P_{dr} \neq P_{mr}$, this indicates that $Sq_i$ changes the reasoning path, and thus $P_{dr}$ cannot be asserted to be a FP. ViFP then analyzes the original path $P_{dr}$ and reconstructs $Sq$. If the TDFM is converted into a non-TDFM in the next round of reasoning, namely, if $(P_{dr}^{+} = P_{mr}^{+}) \ \wedge \ (A^{+} = A^{*}), \text{ then } P_{dr} \notin \text{FP}$, 
where the superscript "+" stands for the next round of reasoning. Otherwise, the instance remains TDFM and is identified as a FP. namely, if $(P_{dr}^{+} = P_{mr}^{+}) \ \wedge \ (A^{+} \neq A^{*}), \text{ then } P_{dr} \in \text{FP}$.
The prompt used in this part is shown as follows:
\begin{figure}[H]
    \centering
    \includegraphics[width=\linewidth]{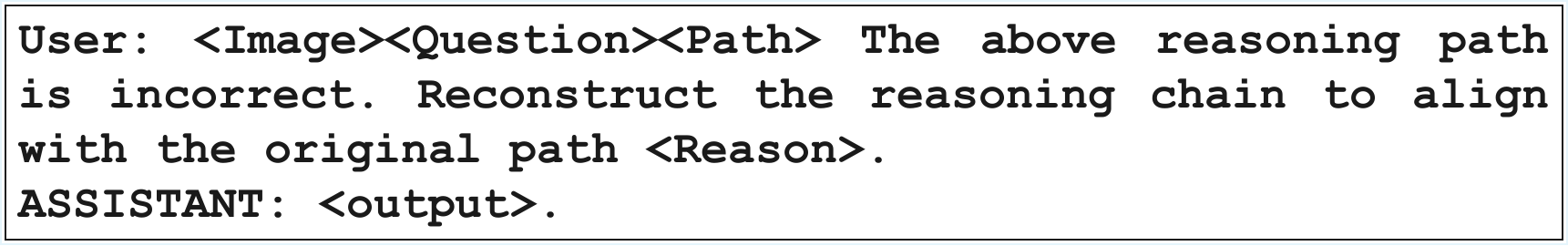}
    \label{FP Detection2}
    \vspace{-10pt}
\end{figure}
The output of the above two parts is a new reasoning chain $Sq$, which is used to summarize the new $Sq_i^{+}$.

\subsection{$QT$ and $Sq_i$ Fine-tuning} 
ViFP is a training-free method but requires fine-tuning of $QT$ and $Sq_i$. Based on the new reasoning chain $Sq$, the prompt used to summarize $Sq_i^{+}$ is shown as follows:
\begin{figure}[H]
    \centering
    \includegraphics[width=\linewidth]{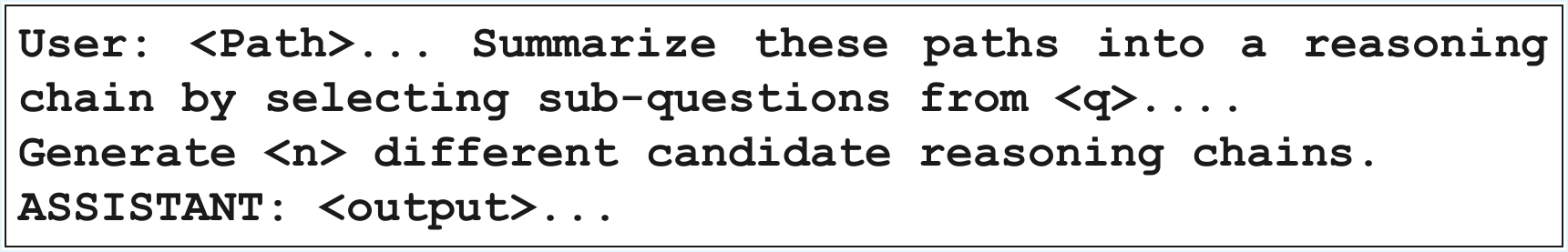}
    \label{Summarize Sq}
    \vspace{-10pt}
\end{figure}
The output consists of n candidate reasoning chains generated by the VLM, and ViFP evaluates them using information gain (IG) to enhance the adaptability of reasoning templates:
\begin{equation}
    IG(Q \mid Sq_{ij}) = H(Q) - H(Q \mid A_j),
\end{equation}
where $H(Q)$ stands for the uncertainty (entropy) of the answer to the question $Q$, the subscript "j" stands for the j-th candidate reasoning chain, and $H(Q \mid A_j)$ denotes the conditional entropy of answering $Q$ given the answers $A_j$. The calculation formula for $H$ is:
\begin{equation}
    H(Q) = -\sum_{i=1}^{n} p(a) \log p(a),
\end{equation}
where $p(a)$ denotes the confidence score of option $a$.

ViFP selects the candidate reasoning chain that maximizes the $IG$ as the final $Sq_i^{+}$:
\begin{equation}
Sq_i^{+}=\arg\max_{Sq_{ij}}\,\frac{1}{n}IG_{total}(Q\mid Sq_{ij}),
\end{equation}

In the next round of reasoning, VLM generates the new output $O_{mr}^{+}$ based on the $Sq_i^{+}$. When $QT = QT_i$, if the corresponding reasoning accuracy increases, then $Sq_i\leftarrow Sq_i^{+}$. If the reasoning accuracy decreases, a new question type $QT_{new}$ is created and $Sq_{new}\leftarrow Sq_i^{+}$, while $Sq_i$ reverts to $Sq_i^{-}$ (the superscript "-" stands for the previous round of reasoning). 

\begin{table*}[htbp]
  \centering 
  \caption{Description of Question Types}
    \begin{tabular}{c|l}
      \toprule
      Question Type & \multicolumn{1}{c}{Description} \\
      \midrule
      Object Localization and Recognition (OLR) & Locate the object in the image and identify what it is. \\
      Temporal Reasoning (TR) & Determine the season, era, or time of day depicted in the image. \\
      Geolocation (GL) & Identify the environment, location, region, or country shown in the image. \\
      Analogical Reasoning (AR) & Compare the consistency between the options and the object in the question. \\
      Functional Reasoning (FR) & Locate and recognize the object, then analyze its function or use. \\
      Intentional Reasoning (IR) & Interpret the object’s next action or determine whether a certain action can be performed. \\
      State Perception (SP) & Select the most appropriate adjective from the options to describe the object's state. \\
      Causal Reasoning (CR) & Explain phenomena or infer causality. \\
      Action Perception (AP) & Understand and analyze the actions of objects in the image. \\
      Spatial Relationship (SR) & Judge positional relationships and estimate distances between objects. \\
      Commonsense Reasoning (COR) & Apply real-world commonsense knowledge to interpret the object's behavior. \\
      \bottomrule
    \end{tabular}
  \label{tab:QT}
\end{table*}

\begin{table*}[htbp]
  \centering
  \caption{Description of Sub-Questions}
  \begin{tabular}{c|l}
    \toprule
    Sub-question & \multicolumn{1}{c}{Description} \\
    \midrule
    Object Discovery (od) & How many objects do you need to focus on according to the image and question?  \\
    Existence Verification (ev) & Do these objects exist in the image? \\
    Object Localization (ol) & Briefly describe their/its location based on the image. \\
    Characteristic Description (cd) & Briefly describe their/its characteristics based on the image. \\
    Scene Description (sd) & Briefly describe the scene based on the image. \\
    Relationship Description (rd) & Briefly describe their relationships based on the image. \\
    Knowledge Retrieval (kr) & Do you need any knowledge to answer this question? \\
    Spatial Relationships Description (srd) & Briefly describe their spatial relationships based on the image. \\
    Temporal Information Discovery (tid) & Is there any object in the image that indicates the time? \\
    Spatial Information Discovery (sid) & Is there any object in the image that indicates the location? \\
    \bottomrule
  \end{tabular}
  \label{tab:SQ}
\end{table*}

\begin{table}[!htbp]
    \centering
    \caption{Reasoning Chain}
    \label{table:Chain}
    \begin{tabularx}{\columnwidth}{
        >{\centering\arraybackslash}p{0.2\columnwidth} |
        >{\centering\arraybackslash}p{0.45\columnwidth}       |              
        >{\centering\arraybackslash}X
    }
    \toprule
    Question Type & \multicolumn{1}{c|}{Reasoning Chain} & \multicolumn{1}{c}{Confidence}\\
    \midrule
      OLR & od $\to$ ev $\to$ ol & 0.908\\
      TR & tid $\to$ od $\to$ cd & 0.895\\
      GL & sid $\to$ od $\to$ kr & 0.960\\
      AR & od $\to$ cd $\to$ sd & 0.920\\
      FR & od $\to$ sd & 0.916\\
      IR & od $\to$ rd $\to$ sd & 0.896\\
      SP & od $\to$ cd & 0.742\\
      CR & od $\to$ sd $\to$ kr & 0.957\\
      AP & od $\to$ ol $\to$ cd $\to$ sd & 0.865\\
      SR & od $\to$ ol $\to$ rd & 0.750\\
      COR & od $\to$ ol $\to$ cd $\to$ kr & 0.783\\
    \bottomrule
    \end{tabularx}
\end{table}

\subsection{Fine-tuning Termination and Outcome}
If $QT_{new}$ is generated, all instances are reclassified. If the new type covers at least 50\% of the incorrect instances from the previous round, it is retained. Otherwise, it is deleted. When ViFP no longer generates any $QT_{new}$, the fine-tuning process stops.

As shown in Tables \ref{tab:QT}, \ref{tab:SQ} and \ref{table:Chain}, after multiple rounds of fine-tuning, ViFP generated 10 sub-questions, 11 question types, each type containing 2 to 4 sub-questions. In Table \ref{table:Chain}, the confidence is calculated as:
\begin{equation}
\text{confidence} = \frac{\text{support}}{\phi},
\end{equation}
where $\text{support}$ denotes the number of instances for which the final answer is correct when the VLM follows $Sq_i$, and $\phi$ stands for the number of questions whose type is $QT_i$.

\section{Experiments}
\subsection{Dataset} 
We conduct our experiments on three VQA datasets: A-OKVQA \cite{schwenk_-okvqa_2022}, OK-VQA \cite{marino_ok-vqa_2019} and FVQA \cite{wang_fvqa_2018}—to evaluate ViFP’s generalizability on real-world VQA tasks. A-OKVQA provides multiple-choice options, while OK-VQA and FVQA are open-ended, with each question having multiple human-annotated answers. The A-OKVQA training set contains 17,056 instances, 1,145 instances in the validation set, and 6,702 instances in the test set. The OK-VQA test set contains 5,046 instances, and the FVQA test set contains 2,778 instances. We fine-tune ViFP on the A-OKVQA training set, as this dataset provides answer options. This setup facilitates the filtering of TDFMs and the subsequent detection of FPs.

\subsection{Baseline} 
We first evaluated the performance of ViFP on three mainstream closed-source VLMs: GPT-4o, Gemini-2.5-flash and Grok-4 in order to evaluate ViFP's cross-model performance. Subsequently, we conducted a comparative analysis with 7 open-source state-of-the-art trainable models: PromptCap \cite{hu_promptcap_2023}, ZS-F-VQA \cite{chen_zero-shot_2021}, DEDR \cite{salemi_symmetric_2023}, MM-Reasoner \cite{khademi_mm-reasoner_2023},SMoLA-PaLI-X \cite{wu_omni-smola_2024} MiniCPM-V-2.6 \cite{yao_minicpm-v_2024} and Prophet++ \cite{yu_prophet_2025}. To intuitively evaluate the effectiveness of the reasoning chain provided by ViFP, we compare ViFP with two CoT approaches—“Let’s think step by step” and LLaVA-CoT in the ablation study.

\subsection{Implementation Detail} 
GPT-4o were used to fine-tune ViFP on the A-OKVQA's training set. We classified all instances into four initial question types and sampled 2,000 instances from the training set. These instances were used by the VLM to generate and refine $QT_i$, $Sq_i$ and so on. During the process, the numbers of $QT$ were 4, 6, 9, 11, 11, while the accuracy plateaued at 90.7\%, after which the fine-tuning process was terminated.

All experiments were conducted on a Linux workstation equipped with a single NVIDIA A800 GPU(80 GB) and Python 3.10. The fine-tuning process on the training set took approximately 24 hours. The total cost was estimated to be under \$ 75 for fine-tuning and under \$100 in total (including fine-tuning and evaluation).

\subsection{Value of FP Correction} 
The Value of FP Correction(VoC) is a novel metric that quantifies how FP correction enhances both reasoning accuracy and reliability in ViFP. Existing VQA evaluation metrics mainly focus on accuracy, which may overlook instances of hallucinated or flawed reasoning paths that lead to correct answers.

Unlike conventional accuracy or path matching metrics, VoC integrates three essential components: (1) The improvement in accuracy achieved through multi-step reasoning; (2) The absolute accuracy of reasoning; (3) The capacity to reduce FPs is indirectly estimated via the true negative (TN) rate, which ensures that fewer FPs lead to a higher VoC.

VoC is defined as
\begin{equation}
    \text{VoC} = (P - Q) \cdot P \cdot \left( \frac{TN}{FP + TN} \right) ,
\end{equation}
where $P$ and $Q$ denote the accuracies of multi-step and direct reasoning, respectively, $FP$ and $TN$ denote the counts of false positives and true negatives. The final term represents the TN rate, namely the proportion of negative instances correctly predicted as negative. 

The partial derivative of VoC with respect to $P$ is:
\begin{equation}
\frac{\partial \text{VoC}}{\partial P} = (2P - Q)\cdot\left( \frac{TN}{FP + TN} \right)
\end{equation}
Note that in the region where \( P > \frac{Q}{2} \), VoC increases with $P$. However, when \( P < Q \), VoC becomes negative, indicating that the introduction of new reasoning chain reduces the reasoning reliability. Therefore, in the desirable region where \( P > Q \), a higher VoC value reflects a greater benefit from multi-step reasoning.

The partial derivative of VoC with respect to $FP$ is:
\begin{equation}
\frac{\partial \text{VoC}}{\partial FP} = \frac{ (Q - P) \cdot P \cdot TN }{(FP + TN)^2}
\end{equation}
Note that in the region where \( P > Q \), VoC increases as the number of FPs decreases, highlighting the importance of minimizing FPs in improving reasoning reliability.

\begin{figure}[!t]
    \centering
    \includegraphics[width=0.8\linewidth]{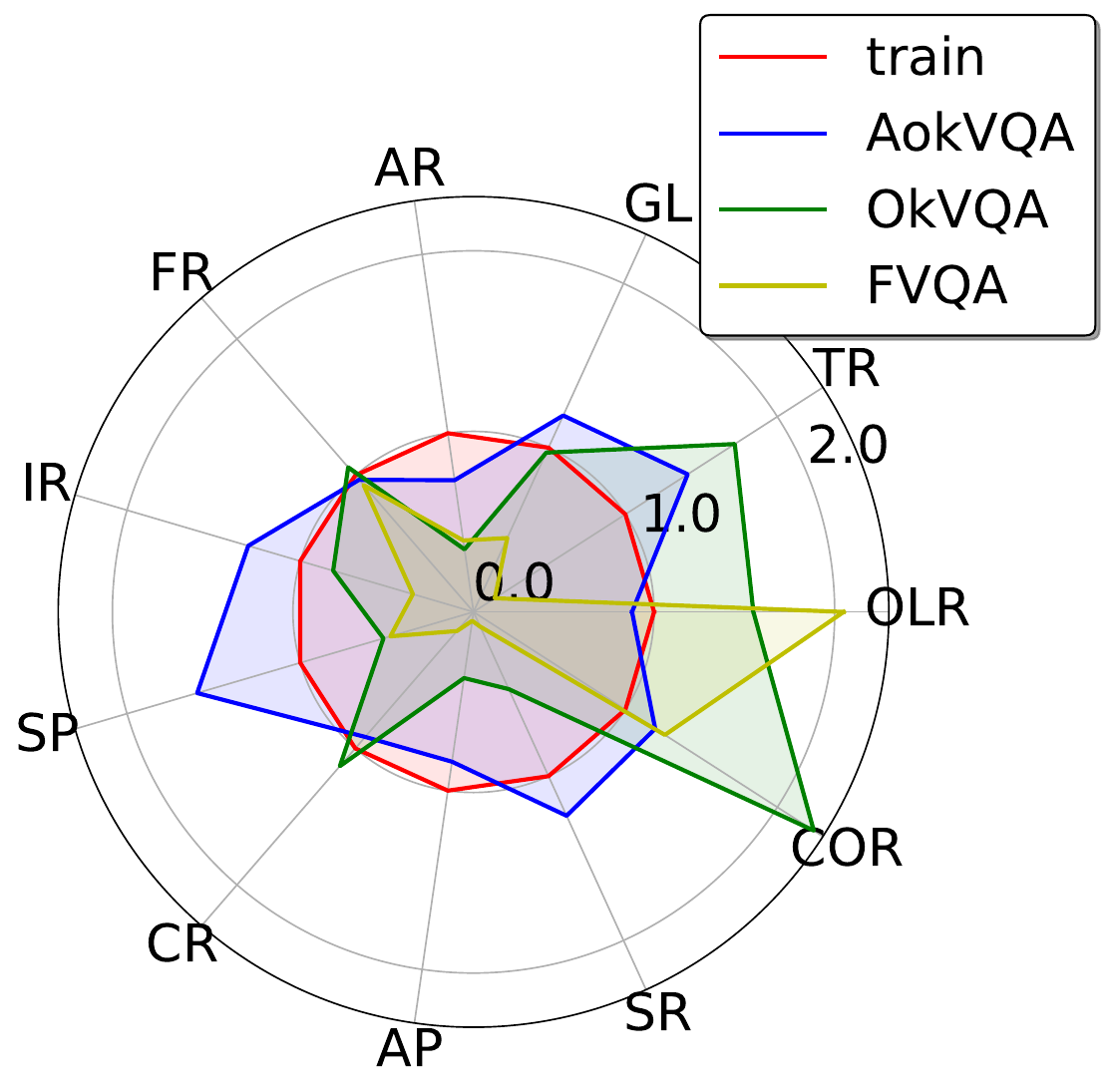}
    \caption{Data distribution of question types on datasets}
    \label{Fig5}
\end{figure}

\begin{table*}[b]
\centering
\caption{Difficulty and Consistency Score$V(t)$}
\begin{tabular}{c|ccccccccccc}
\toprule
 & \textbf{OLR} & \textbf{TR} & \textbf{GL} & \textbf{AR} & \textbf{FR} & \textbf{IR} & \textbf{SP} & \textbf{CR} & \textbf{AP} & \textbf{SR} & \textbf{COR} \\
\midrule
$\mathrm{Dif}(QT_i, D_{A-OKVQA})$ & 0.53 & -0.33 & -4.82 & 0.17 & -0.90 & 1.79 & 3.30 & -1.11 & -1.41 & 3.44 & 2.44 \\
$\mathrm{Dif}(QT_i, D_{OK-VQA})$ & -2.36 & -0.24 & -0.11 & -0.13 & 0.52 & 0.56 & -0.15 & 0.68 & -0.11 & 0.15 & -0.28 \\
$\mathrm{Dif}(QT_i, D_{FVQA})$ & -10.93 & 0.42 & 4.01 & -0.39 & 2.67 & 1.76 & 0.32 & 0.09 & -0.06 & 0.00 & 2.60 \\
\midrule
$\overline{\mathrm{Dif}}(QT_i)$ & -4.26 & -0.05 & -0.32 & -0.11 & 0.05 & 1.14 & 1.16 & -0.11 & -0.53 & 1.20 & 1.59 \\
$\mathrm{V}(QT_i)$ & 23.68 & \textbf{0.11} & 11.99 & \textbf{0.05} & 5.52 & \textbf{0.14} & \textbf{2.83} & \textbf{3.32} & \textbf{0.39} & \textbf{2.54} & \textbf{1.55} \\
\bottomrule
\end{tabular}
\label{tab:difficulty}
\end{table*}
\subsection{Evaluation Strategy}

(1) \textbf{Generalizability} is evaluated by measuring the difficulty consistency score of each question type across multiple real-world VQA datasets. 

\( E_{QT_i,D_j}^{(m)} \) denotes the number of incorrect answer instances of type \( QT_i \) made by VLM \( m \) on dataset \( D_j \), \( E_{D_j}^{(m)} \) denotes the total number of incorrect answer instances, \( N_{QT_i,D_j} \) denotes the total number of questions of type \( QT_i \) in \( D_j \), with \( N_{D_j} \) representing the total number of all questions in \( D_j \). 

The relative difficulty of type \( QT_i \) on dataset \( D_j \) for VLM \( m \) is defined as the difference between its error proportion and its occurrence proportion:
\begin{equation}
\mathrm{Dif}_{QT_i,D_j}^{(m)} = \frac{E_{QT_i,D_j}^{(m)}}{E_{D_j}^{(m)}} - \frac{N_{QT_i,D_j}}{N_{D_j}} .
\end{equation}

The dataset-level difficulty of question type \( QT_i \) is computed by averaging across all \( M \) VLMs (here \( M = 3 \)):
\begin{equation}
\mathrm{Dif}(QT_i, D_j) = \frac{1}{M} \sum_{m=1}^{M} \mathrm{Dif}_{QT_i,D_j}^{(m)} .
\end{equation}

For each $QT$, \( N \) datasets correspond to \( N \) difficulty scores. 
We then measure the difficulty consistency score \( \mathrm{V}(QT_i) \):
\begin{equation}
\mathrm{V}(QT_i) = \frac{1}{N}\sum_{i=1}^{N}\left(\mathrm{Dif}(QT_i, D_j) - \overline{\mathrm{Dif}}(QT_i)\right)^2 ,
\end{equation}
where the mean difficulty across datasets is
\begin{equation}
\overline{\mathrm{Dif}}(QT_i) = \frac{1}{N}\sum_{i=1}^{N}\mathrm{Dif}(QT_i, D_j) .
\end{equation}

A smaller \( \mathrm{V}(QT_i) \) indicates that the type $QT_i$ exhibits similar difficulty levels across different datasets, and the performance of ViFP does not drop sharply when applied to these datasets.

\begin{figure}[!t]
    \centering
    \includegraphics[width=0.9\linewidth]{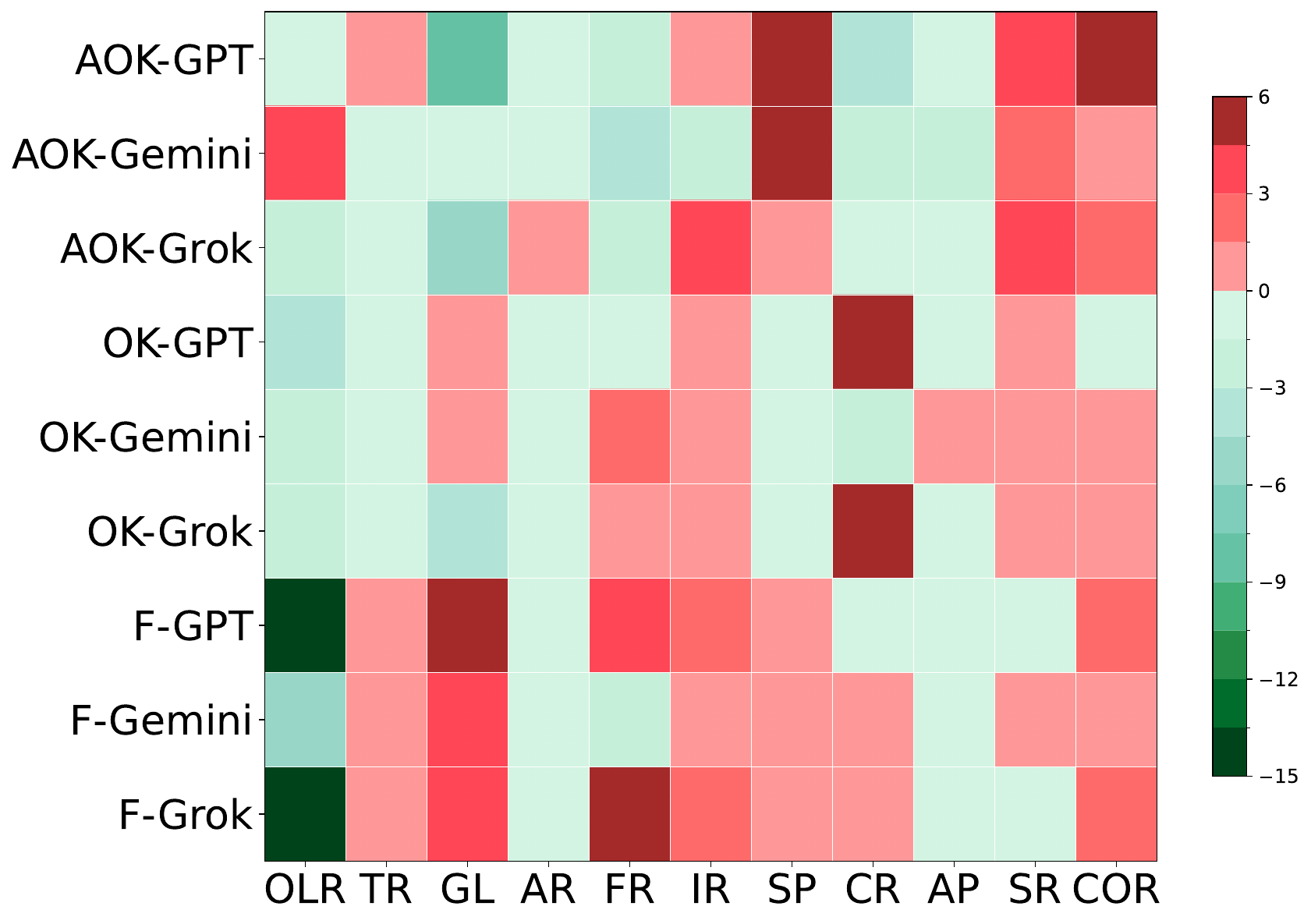}
    \caption{Difficulty of different question types on datasets}
    \label{Fig6}
\end{figure}

(2) \textbf{Reasoning accuracy} is assessed on three real-world VQA datasets. A-OKVQA provides four options with only one correct answer. The standard accuracy is adopted as the evaluation metric. OK-VQA and FVQA are open-ended, the top-k matching metric is used as the evaluation metric: the VLM outputs multiple candidate answers, and Top-1 is defined as the probability that the first candidate matches the ground truth. Top-3 is defined as the probability that the ground truth appears among the top three candidates. In order to align with baselines, FVQA is evaluated using both Top-1 and Top-3, whereas OK-VQA only uses Top-1.

(3) \textbf{Reasoning reliability} is evaluated using three indicators: VoC, FP, and TDFM counts. FP and TDFM counts track the reduction of FP instances and quantify the contribution of newly generated reasoning chains by ViFP to improvements in accuracy.



\begin{table}[htbp]
\centering
\caption{Comparison between ViFP-based VLMs and trainable models.}
\setlength{\tabcolsep}{1mm}
\small
\renewcommand{\arraystretch}{1.1}
\begin{tabular}{l|cccc}
\toprule
\textbf{Model} &\textbf{A-OKVQA} &\textbf{OK-VQA} & \textbf{FVQA}\textsubscript{Top-1} & \textbf{FVQA}\textsubscript{Top-3}  \\
\midrule
\textbf{\textit{Trainable}}\\
PromptCap                & 73.2     & 60.4         & -       & -    \\
ZS-F-VQA                 & -        & -            & 58.3    & 75.2 \\
DEDR                     & -        & 51.0         & 61.8    & -    \\
MM-Reasoner              & -        & 60.8         & 61.1    & -    \\
SMoLA-PaLI-X             & 84.1     & 62.4         & -       & -    \\
MiniCPM-V-2.6            & 86.4     & -            & -       & -    \\
Prophet++                & 87.7     & \textbf{65.7}& -       & -    \\
\midrule
\textbf{\textit{Train-free}}\\
Grok-4             & 78.1     & 42.3     & 64.4     & 69.2          \\
Gemini-2.5         & 86.6     & 46.3     & 64.8     & \textbf{75.2} \\
GPT-4o             & 88.8     & 49.0     & 69.6     & 71.7          \\
\midrule
Grok-4+ViFP             & 80.0          & 48.2     & 66.9           & 70.4   \\
Gemini-2.5+ViFP         & \textbf{92.0} & 40.9     & 61.9           & 66.3   \\
GPT-4o+ViFP             & 90.7          & 55.2     & \textbf{70.3}  & 73.8   \\
\bottomrule
\end{tabular}
\label{tab:results}
\end{table}


\subsection{Data distribution}
As shown in Fig. \ref{Fig5}, we first summarize the distribution of different question types across the three datasets. Second, we compute the relative difficulty and difficulty consistency $V$ of each question type. As shown in Fig. \ref{Fig6} and Table \ref{tab:difficulty}, Eight question types exhibit a relatively low $V(QT_i)$ value (below 3.5). Only OLR, GL and FR show slightly higher $V(QT_i)$, which should be given more attention in ViFP's applications. However, as their $Dif$ values are relatively low, they don't significantly affect the generalizability of ViFP. As shown in Table \ref{tab:results}, this is further corroborated by the fact that the accuracy of most VLM+ViFP combinations improves accordingly.

Based on the above information, we can identify which question types the VLM handles more effectively, providing useful guidance for further refining ViFP. Furthermore, these findings also offer recommendations for applying closed-source VLMs and for performing targeted supervised fine-tuning. First, the data distribution for the validation set of A-OKVQA is most similar to the training set, and SP, SR, and COR are relatively difficult, making them key areas to focus on in the next reasoning (refine $QT$ and $Sq_i$). Grok is more proficient at answering these questions. OK-VQA has more COR and TR, with CR being the most difficult. GPT is better at answering. FVQA has a large number of OLR questions, but they are relatively easy. To further fine-tune ViFP on FVQA, more attention should be given to GL, FR and COR.

\subsection{Visual Reasoning} 
\begin{figure}[t]
    \centering
    \includegraphics[width=0.9\linewidth]{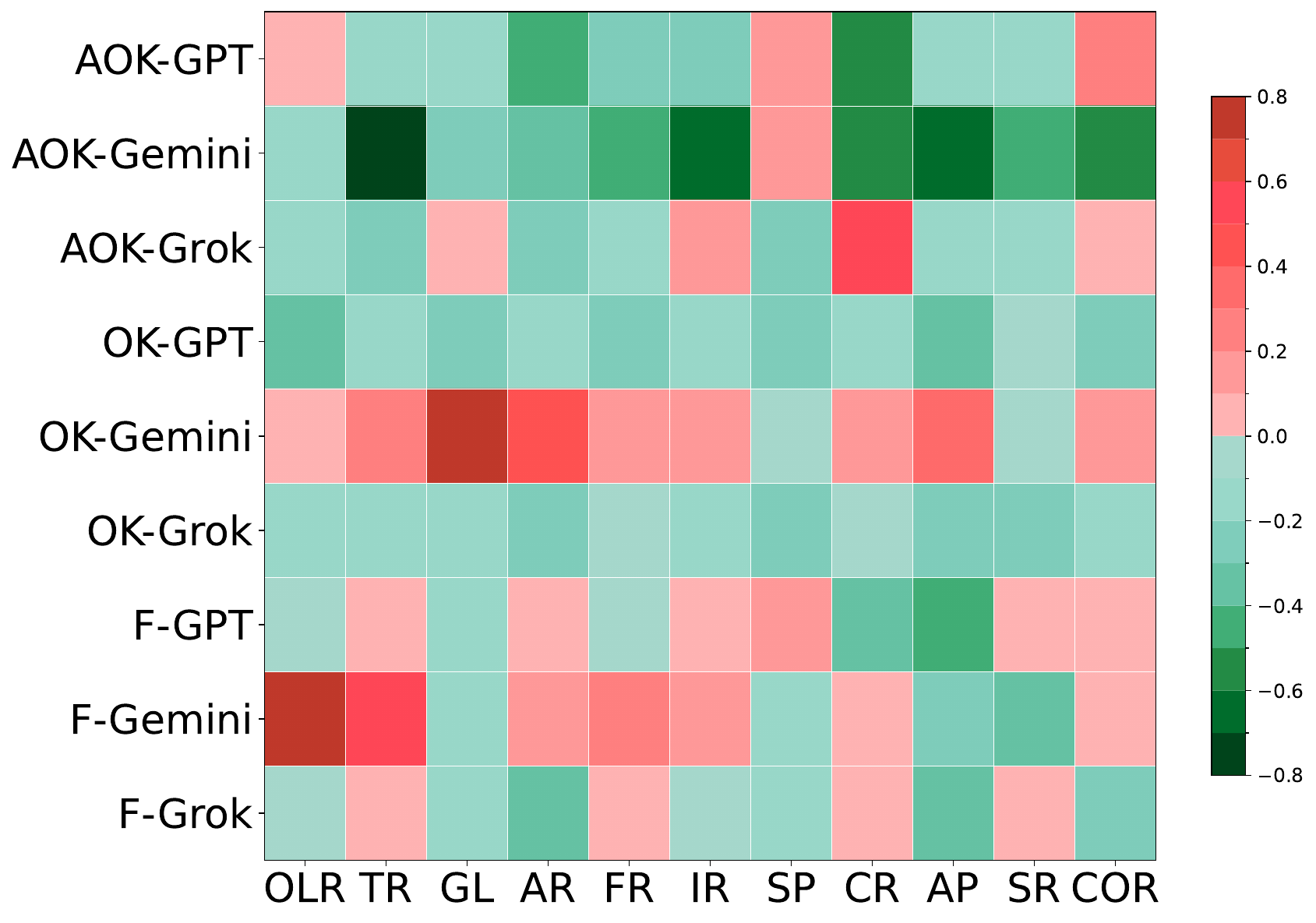}
    \caption{Improvement of ViFP-base VLMs}
    \label{Fig7}
\end{figure}

ViFP is evaluated on Grok-4, Gemini-2.5, and GPT-4o. Table \ref{tab:results} and Fig. \ref{Fig7} show the reasoning accuracy and improvements on three real-world VQA datasets.
Table \ref{Fig7} reveals that ViFP significantly enhances the reasoning accuracy of closed-source VLMs. As shown in Fig. \ref{Fig7}, ViFP can help VLM to handle most types of questions, especially GPT and Gemini on A-OKVQA, GPT and Grok on OK-VQA, Grok on FVQA (green represents improvement, red represents decline).

The only exception is ViFP-based Gemini on OK-VQA and FVQA. This phenomenon occurs primarily because accuracy on open-ended VQA datasets is greatly influenced by the model’s parameter scale and response preferences. The parameter scale of Gemini-2.5-flash is smaller than that of GPT-4o and Grok-4. Its logical reasoning ability is relatively weak, making it prone to over-interpretation or generating excessive irrelevant reasoning in multi-step reasoning processes. Although it generates many approximate answers, its accuracy decreases because they do not completely match the ground truth. In addition, when provided with longer reasoning cues, Gemini also tends to produce longer responses. We counted the number of instances in which the VLM's answer contained more than one word while the correct answer consisted of only one word. Gemini-2.5 had 1370 such instances, Grok-4 had 683, and GPT-4o had the fewest, with only 171. Therefore, if the number of words in the answer can be specified accurately, these VLMs, particularly Gemini-2.5, are likely to achieve significantly higher accuracy on open-ended VQA datasets.

Finally, we analyzed the incorrect answers in Gemini-2.5+ViFP and summarized them as follows:

(1) Morphological disagreement and semantic granularity errors: such as Gemini-2.5+ViFP’s answers being "vases"($A^{*}$: vase) and "Kitchen island"($A^{*}$: island); 

(2) Over-interpreted and speculation: such as the question is "Who leaves a toilet like this?" ($A^{*}$: man), the answer of Gemini-2.5 is "Everyone" and the answer of Gemini-2.5+ViFP is "Someone untidy", because "toilet... has its lid and seat up", ViFP helps VLM to focus the new clue: toilet seat, and the multi-step reasoning process is closer to the ground truth than direct answering. However, since the VLM is unaware that the ground truth is man/woman, it overinterprets as “untidy”;

(3) Category-level mismatches: such as the question "What type of plane is that?" ($A^{*}$: commercial), where Gemini-2.5+ViFP’s answer is "Boeing 737-800";

(4) Preference for internal knowledge: such as the question "What does this grow from?" ($A^{*}$: ground), where Gemini-2.5’s answer "From a bulb".


\subsection{Ablation Study} 
Ablation studies are conducted to verify the effectiveness of the components in ViFP. Moreover, VoC is introduced to intuitively show the contribution of the FP detection module.

\begin{table}[t]
\setlength{\tabcolsep}{1.5mm}
\caption{Ablation study results on A-OKVQA, “VoC-base” indicates that all ablation studies take this result as the baseline.}
\centering
\small
\begin{tabular}{l|cccc}
\toprule
\textbf{Model}                  & \textbf{Acc}  & \textbf{VoC}   & \textbf{FP}  & \textbf{TDFM}\\
\midrule
GPT-4o (VoC-base)               & 88.8          & 0              & --           & --\\
GPT-4o+Let's think step by step & 89.8          & 0.83           & 5            & 23\\
GPT-4o+LLaVA-CoT                & 89.3          & 0.36           & 11           & 34\\
\midrule
GPT-4o+ViFP (QT=6)              & 85.1          & -2.62          & 28           & 74\\
GPT-4o+ViFP (QT=9)              & 90.5          & 1.32           & 18           & 28\\
GPT-4o+ViFP (QT=11)             & 90.7          & 1.48           & 16           & 19\\
Gemini-2.5+ViFP (QT=11)         & \textbf{92.0} & \textbf{2.47}  & 22           & 36\\
\bottomrule
\end{tabular}
\label{tab:Ablation-study}
\end{table}

\subsubsection{Impact of Question Type Granularity on Reasoning}
As shown in Table \ref{tab:Ablation-study}, we compare the accuracy under different $QT$s using GPT-4o. In the early stage, when the $QT$s were still coarse (QT = 6), a significant proportion of questions were ambiguously classified, the $Sq_i$ failed to provide effective reasoning guidance and thus negatively affected the reasoning accuracy. As classification became more refined (QT = 11), the accuracy gradually improved, indicating that the VLM can arrive at the correct answer by refining its $Sq_i$.
    
\subsubsection{ViFP Compared to Generic CoT Method}
We further compare the reasoning chains generated by ViFP with generic CoT prompts. Table \ref{tab:Ablation-study} shows that the generic CoT prompts provide limited benefits to the VLM. Their chains lack specificity and are ineffective in guiding the VLM to focus on the key elements necessary for answering the question. When VLM does not know the explicit reasoning path, it tends to describe the image comprehensively by generating image captions, rather than trying to focus on the key points in the image according to the question, potentially resulting in a FP or accumulation of errors. In contrast, the reasoning chains generated by ViFP explicitly guide the VLM to focus on the key visual information in the images and offer the reasoning paths, significantly improving answer accuracy and reasoning reliability.

\subsubsection{Validating Reasoning Reliability with VoC}
VoC is calculated based on the results of direct reasoning and multi-step reasoning. Multi-step reasoning places greater demands on the VLM's contextual understanding and cognitive abilities. Therefore, closed-source VLMs are used to compute VoC, FP and TDFM in order to better evaluate the improvement in reasoning reliability.

As $QT$ is refined, the reliability of multi-step reasoning improves significantly. First, the number of TDFMs decreases steadily, indicating that errors caused by the reasoning chains produced by ViFP are being reduced. Second, the number of FPs decreases while the FP/TDFM ratio increases. This suggests that more TDFMs are converted into non-TDFMs, as more FPs are corrected to true positives. In addition, VoC usually increases with accuracy, but doesn’t always decrease as FP decreases. This is because FP only reflects the number of FPs that can be detected, can not fully account for all latent incorrect reasoning. However, with similar accuracy, a lower number of FPs indicates a higher VoC. This suggests that multi-step reasoning in this round has a higher reasoning reliability.

\subsubsection{Significance of negative VoC}
It is worth noting that when QT = 6, VoC is negative (-2.62). This is because the reasoning chains actually weaken the VLM's reasoning ability, resulting in the generation of unreasonable paths. This observation is of great significance, as it quantitatively demonstrates that a poorly designed reasoning chain not only fails to help but also interferes with the VLM’s normal reasoning process, highlighting the necessity of refined question classification and reasoning chain optimization. As the $QT$ and $Sq_i$ become more refined, the VLM achieves improvements in reasoning accuracy and reliability simultaneously.

\begin{figure}[t]
    \centering
    \includegraphics[width=0.8\linewidth]{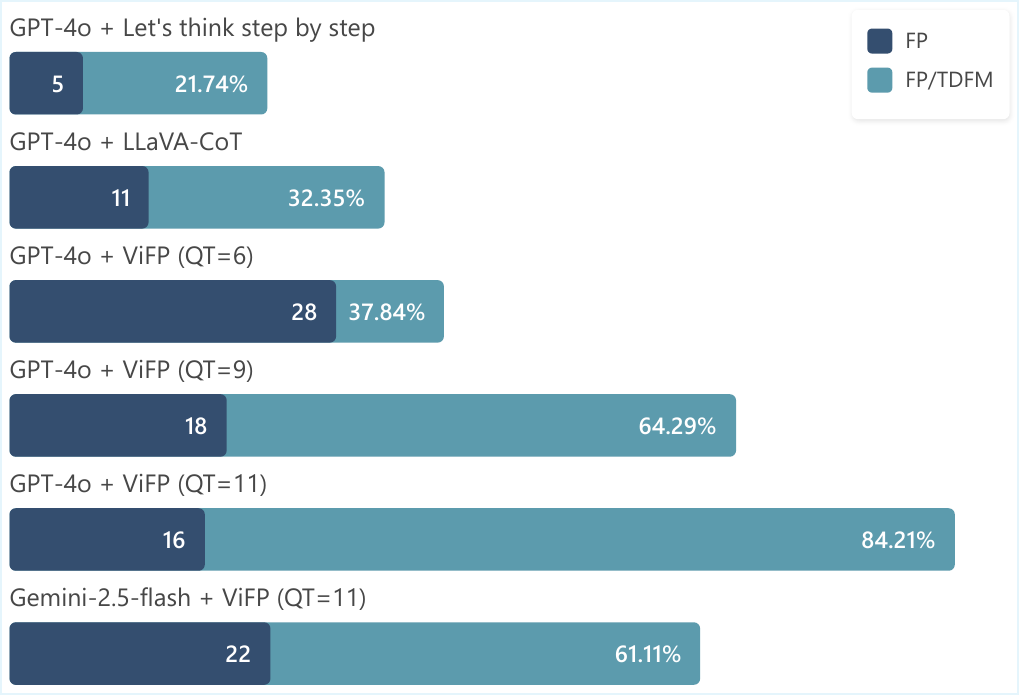}
    \caption{The variation of FPs and FP/TDFM.}
    \label{Fig8}
\end{figure}

\subsection{Case Study and Visualization of Accuracy Improvement Mechanisms}
\begin{figure*}[htbp]
    \centering
    \includegraphics[width=\linewidth]{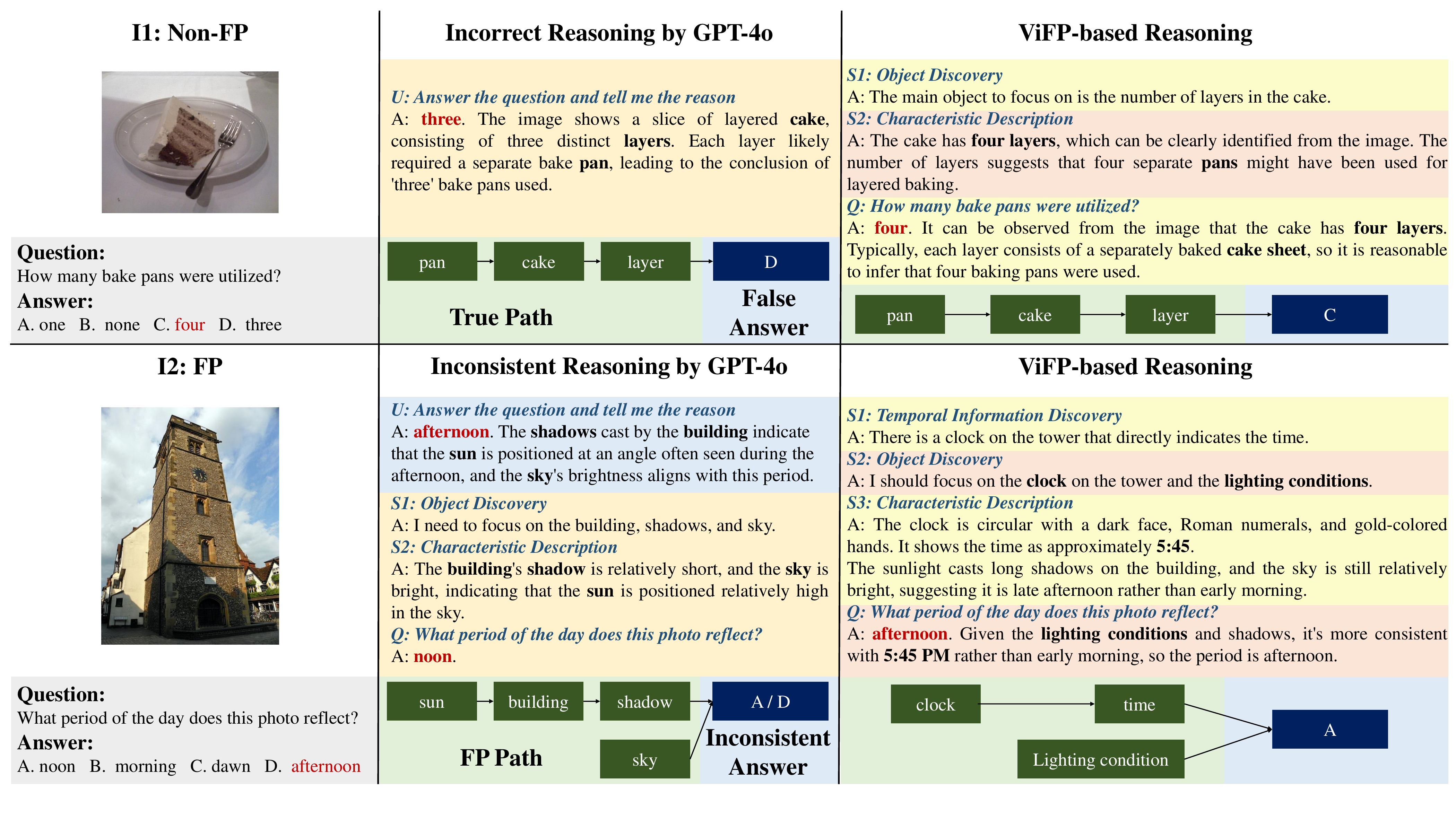}
    \caption{Two instances of how ViFP works: In instance 1 (I1), ViFP guides the VLM to focus on more details in the image by providing an effective reasoning chain. In I2, ViFP corrects unreliable reasoning through FP detection, and guides VLM to focus on the key information: the "clock".}
    \label{FigInstances}
\end{figure*}

Based on the comparison with existing CoT methods, we summarize that ViFP mainly serves two functions. It improves VQA accuracy by providing effective and explicit reasoning chain, and enhances reasoning stability and reliability by correcting unreliable reasoning chains. Fig. \ref{FigInstances} illustrates the two improvement instances.

\section{Conclusion}
In this paper, we propose ViFP, which can be applied directly to VLMs with no need for additional training. ViFP enables VLMs to detect FPs independently. Unlike traditional cross-model verification methods, ViFP uncovers a broader range of potential FPs. Furthermore, ViFP introduces a visual reasoning optimization method based on feedback from FPs, which adjusts question types and reasoning chains. This guides the VLM towards more reliable reasoning paths and more accurate answers. We also introduce a novel metric called VoC to evaluate the value of FP correction in multi-step reasoning. VoC evaluates both the accuracy of reasoning answers and the soundness of reasoning paths.

In future work, we plan to incorporate reinforcement learning into the false alarm detection process in order to better calculate the rationality score of inference paths and further strengthen ViFP's FP detection capabilities. This will provide more entry points for improving visual reasoning reliability.

\bibliographystyle{IEEEtran}
\bibliography{TMM}

\vfill

\end{document}